# A Comprehensive Study on NLP Data Augmentation for Hate Speech Detection: Legacy Methods, BERT, and LLMs⋆

Md Saroar **Jahan**[a,*], Mourad **Oussalah**[a], Djamila Romaissa **Beddia**[a], Jhuma kabir **Mim**[b] and Nabil **Arhab**[a]

[a]*University of Oulu, CMVS, BP 4500, 90014, Finland*
[b]*LUT Univerity, Computational Engineering, 53850, Finland*

## ABSTRACT

The surge of interest in data augmentation within the realm of natural language processing (NLP) has been driven by the need to address challenges posed by hate speech domains, the dynamic nature of social media vocabulary, and the demands for large-scale neural networks requiring extensive training data. However, the prevalent use of lexical substitution in data augmentation has raised concerns, as it may inadvertently alter the intended meaning, thereby impacting the efficacy of supervised machine learning models. In pursuit of suitable data augmentation methods, this study explores both established legacy approaches and contemporary practices such as Large Language Models (LLM), including GPT in Hate Speech detection. Additionally, we propose an optimized utilization of BERT-based encoder models with contextual cosine similarity filtration, exposing significant limitations in prior synonym substitution methods. Our comparative analysis encompasses five popular augmentation techniques: WordNet and Fast-Text synonym replacement, Back-translation, BERT-mask contextual augmentation, and LLM. Our analysis across five benchmarked datasets revealed that while traditional methods like back-translation show low label alteration rates (0.3-1.5%), and BERT-based contextual synonym replacement offers sentence diversity but at the cost of higher label alteration rates (over 6%). Our proposed BERT-based contextual cosine similarity filtration markedly reduced label alteration to just 0.05%, demonstrating its efficacy in 0.7% higher F1 performance. However, augmenting data with GPT-3 not only avoided overfitting with up to sevenfold data increase but also improved embedding space coverage by 15% and classification F1 score by 1.4% over traditional methods, and by 0.8% over our method. These results highlight significant advantages of using LLMs like GPT-3 for data augmentation in NLP, suggesting a substantial leap forward in machine learning model performance, particularly for hate speech detection tasks.

## 1. Introduction

Over the last few years,various machine learning and deep learning-based approaches have been suggested for online hate speech detection and tracking within the information processing research community [33, 15], leveraging recent progress in natural language processing research, including sentiment analysis, sentence categorization, text classification, discourse analysis, causality, among others. Nevertheless, despite progress this progress, several challenges still remain in achieving high success rate in uncovering various forms of hate, especially with the growing web and social media vocabulary, inherent language complexity with subjective interpretation of the meaning and increasing forms of implicit hate that often goes beyond language barriers. So far the dominant trend in online hate detection relies on capitalizing on capabilities of deep-learning models utilizing state-of-the-art pre-trained models fine-tuned with existing hate speech dataset. For instance, Agrawal and Awekar, Agrawal and Awekar compared the performance of a set of machine learning-based models (logistic regression, support vector machine, random forest, naive Bayes) and a set of deep learning-based models (CNN, LSTM, BLSTM, BLSTM with Attention using variety of representation methods for words (bag of character n-gram, bag of word unigram, GloVe embeddings, SSWE embeddings)) in capturing hate on various social media platforms (Formspring, Twitter and Wikipedia) and topics (personal attack, racism, and sexism). Bu and Cho [3] proposed a hybrid deep learning model that combines a character-level CNN model and

---

⋆This work is supported by the European Young-sters Resilience through Serious Games, under the Internal Security Fund-Police action: 823701-ISFP-2017-AG-RAD grant, which is gratefully acknowledged.

*Corresponding author

mjahan18@edu.oulu.fi (M.S. Jahan); Mourad.Oussalah@oulu.fi (M. Oussalah)
ORCID(s): 0000-0001-5734-3328 (M.S. Jahan)

---

word-level Long-term Recurrent Convolutional Networks (LRCN) in a way to capture syntactics and semantics of cyberbullying comments from Kaggle. A character-level CNN model with shortcuts was proposed by [28] for identifying hate from Twitter and Chinese posts. Recently, the rise of BERT and transformer architectures [12] reached a momentum in text classification tasks (38% share of deep-learning models) in the past five years [13], see HateBERT [4] for a state-of-the-art BERT model designed for hate speech detection. Despite the success of these models in several applications, moderate success has been reported when dealing hate speech detection, mainly due to generalization issue or overfitting, in addition to inherent NLP-based approach for detecting various forms of implicit hate. Typically, generalization performance highly depends on the size and quality of the training data and regularization [25]. Preparing a large annotated dataset is very time-consuming, and, to some extent, subjective as well. Instead, automatic data augmentation becomes popular to overcome this limitation, particularly in the areas of computer vision [17] and speech recognition [19]. In this respect, it is interesting to seek the contribution of data augmentation techniques to enrich existing annotated hate dataset in a way that would enhance the performance of the employed deep-learning model. Generally, data augmentation (DA) utilizes human knowledge on existing invariance, empirical rules, or heuristics observed in the original dataset, e.g., "even if a picture is flipped, the class of an object should be unchanged". Though, the usage of data augmentation in NLP is very challenging. It is, for instance, challenging to obtain universal rules for transformations in natural languages that preserve the semantic meaning and scale across domains. In this respect, two methods can be distinguished from previous works. The first one consists of replacing the word with its synonyms in the hope of generating close-meaning sentences as in [20, 36]. For instance, WordNet lexical database [31, 45] has been suggested to guide the synonym replacement task. Another approach uses pre-trained word embeddings such as GloVe, FastText, Sent2Vec, where the nearest neighbor word in the embedding space can be used as a replacement for the selected word in the sentence. Alternatively, word similarity calculation has been suggested in [41]. The second type of approaches relies on inserting noise in the original sentence in a way to not alter its meaning. For instance, Wei and Zou [42] suggested an Easy Data Augmentation (EDA) technique that uses a set of four simple operations: random synonym replacement, random synonym insertion, random word deletion, random swap of words in the sentence. The approach was tested on five benchmark classification tasks and exhibited a systematic improvement, especially when smaller training sample was used. Although the interaction and aggregation between the four operations was not elaborated. To handle such an issue, Kobayashi [20] proposed a word substitution approach that preserves sentence label. This is implemented using a bidirectional RNN language model with a label-conditional architecture, which allows the model to augment sentences without breaking the label-compatibility. However, the approach also incurs high computational cost with respect to its performance gain. Kobayashi [20] proposed the conditional BERT (CBERT) model, which extends BERT [6] masked language modeling (MLM) task by considering class labels to predict the masked tokens. Nevertheless, since the method relies on modifying the BERT model's segment embedding, it cannot be generalized to other pretrained LMs without segment embeddings. In the same spirit, Kumar et al. [22] used BERT-based contextual augmentation for word replacement and achieved a better performance compared to EDA. Other notable techniques include back-translation (BT) [37] and paraphrasing [2] augmentation, which have shown promising results. Nevertheless, the accuracy of the model when applied to cross-domain is often questioned, which may lead to dataset labeled alteration. Furthermore, in recent advancements, Large Language Models (LLM) have emerged as pivotal tools in natural language processing, revolutionizing various tasks such as language understanding, generation, and representation. LLMs, exemplified by models like GPT, capitalize on pre-training on vast datasets to capture intricate language patterns. Their application in text classification tasks has shown promising results, prompting exploration in the domain of data augmentation. However, the specific contribution and effectiveness of LLMs in the context of data augmentation for hate speech detection remain under-explored. This calls for further research to fill in these gaps. The current paper proposes a new optimized use of the BERT model using contextual cosine similarity in an iterative loop at the end of each data augmentation pipeline to filter out the closest augmented sentences compared to the original sentence. On the other hand, considering the potential LLMs, this study investigates the impact of integrating LLMs, such as GPT, into the data augmentation pipeline for enhancing the performance of supervised machine learning models in hate speech detection. For this purpose, five benchmark datasets from different domains have been selected and a set of baseline models have been employed to evaluate the performance of the developed model (s). Two datasets were related to cyberbullying and hate speech. Two others were related to sentiment analysis, and the fifth one to question categorization with six different question types; see, Section 4.1 for the details of this process. We tested four popular DA methods: synonym replacement with FastText and wordnet, BT, and BERT MLM as a baseline models. We compared the result of each method after applying BERT-cosine similarity as a label filtration condition. Our method offers a wider range of substitute words and easy implementation. Furthermore, it reveals some limitations of previous methods and prevents word replacement approaches that were incompatible with the annotated labels of the original sentences. Through the experiment, we demonstrate the proposed contextual filtration improves classifiers' performance and reduces significant label alteration. This research answers the following research questions:

**Q1.** Are all previous augmentation methods capable of sentence augmentation without changing the original labeling?

**Q2.** Does BERT-cosine similarity measure improve the database augmentation quality in terms of less label alteration and machine learning performance?

**Q3.** Does the combination of different DA methods helps enrichment of dataset quality, or does it pose an adverse effect by over-fitting the classifier of having too many augmented sentences?

**Q4.** To what extent does the incorporation of Large Language Models (LLM), such as GPT, in the data augmentation process contribute to the improvement of supervised machine learning models in hate speech detection, considering their language understanding capabilities?

The main contributions of this paper are as follow:

1. We experimented with a novel augmentation scheme guided by the use of BERT contextual embeddings, leveraging Cosine similarity to assess closeness with the original sentence. This approach enhances the reliability of the augmented sentences.

2. We compared our proposed data augmentation approach against the most popular state-of-the-art practices and addressed major flaws of previous techniques that have been overlooked by previous work.

3. We developed a new python library for data augmentation, which is the end product of our experiment, and released it under an open-sourced license for the research community [1]. This library ensures the highest possibility of producing close-meaning sentences, as proved by our experiment.

The paper is structured as follows: Section 2 presents a brief review of related work about previous augmentation techniques in the NLP field. Section 3 highlights the overall methodology. We discuss the experimental results and evaluation in Section 4, model performance and error analysis in Section 5, and finally, we discuss and conclude the paper in Section 6 & 7 and outline some future research directions.

## 2. EXISTING METHODS FOR NLP AUGMENTATION

In this section, we briefly discuss key existing methods and tools for NLP data augmentation that try to use substitute words without altering the meaning of the sentence.

1. *Thesaurus-guided Synonym Substitution*: This method involves the random selection of a word from a sentence and its subsequent replacement with a synonym sourced from a Thesaurus. An example of this practice includes utilizing WordNet database for English language substitutions (as shown in Figure 1). It is a manually curated database with relations between words. Zhang *et al.* [45] used this technique in their paper "Character-level Convolutional Networks for Text Classification". Mueller *et al.* [32] used a similar strategy to generate additional 10K training examples for their sentence similarity model. Further, Wei et al. promoted this technique in their pool of four random data augmentation strategies mentioned in the "Easy Data Augmentation" paper [42].

2. *Word-Embeddings Synonym Substitution*: This technique utilizes pre-trained word embeddings like Word2Vec, GloVe, FastText, and Sent2Vec. It replaces certain words in a sentence with their nearest neighbors in the embedding space. In our study, we leveraged the Wikipedia corpus[2] to generate word vectors using FastText word embeddings (an example of augmentation is illustrated in Figure 2). This method was used by Jiao et al. with GloVe embeddings in their "TinyBert" research paper to enhance the generalizability of their language model for downstream tasks [16]. Likewise, Wang et al. applied it for tweet augmentation in learning a topic model [41].

---

[1] https://pypi.org/project/nlp-augment/
[2] http://mattmahoney.net/dc/enwik9.zip (last accessed Aug.24.2021)

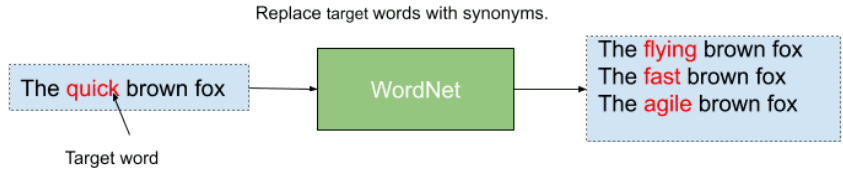

**Figure 1:** Sentence augmentation example using WordNet.

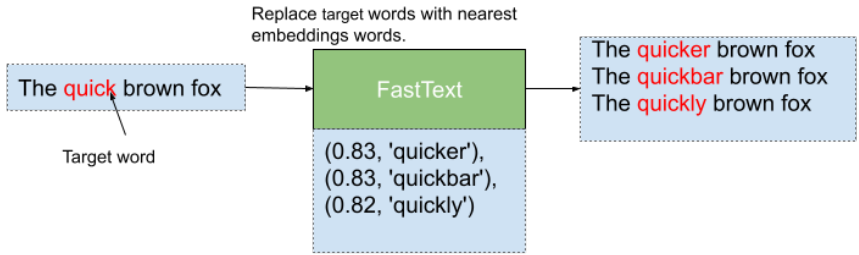

**Figure 2:** Sentence augmentation example using FastText.

3. *Transformer models*: such as BERT [6], ROBERTA [27], and ALBERT [23] have been trained on a large amount of text using a pretext task called "Masked Language Modeling", where the model has to predict masked words based on the context. This can be used to augment some text. For example, by hiding a given token of the sentence, the pre-trained BERT model can be used to predict the masked token. Therefore, variations of original text using such mask prediction can be generated. Compared to previous approaches, the generated text is more grammatically coherent as the model takes context into account when making predictions. However, one limitation of this method is that deciding which part of the text to mask is not inconsequential. Heuristics to decide the mask have to be explored; otherwise, the generated text might not retain the meaning of the original sentence. Garg *et al.* [7] used this idea for generating adversarial examples for text classification.

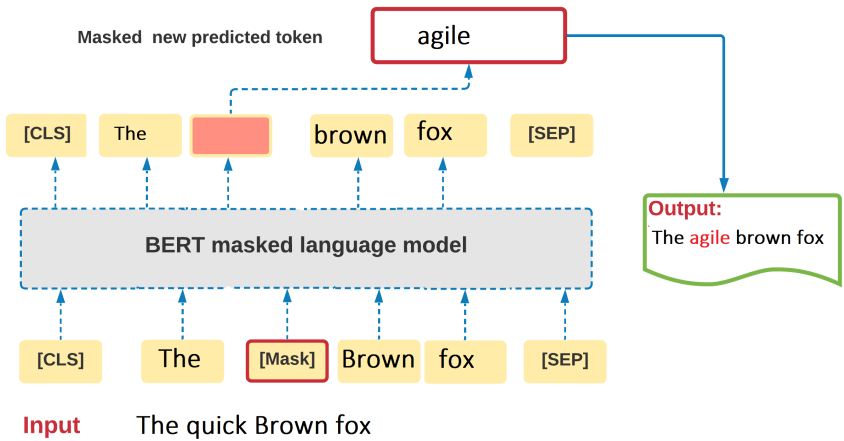

**Figure 3:** Example of replacing a word with a <mask> token.

4. *Back-translation (BT)*: In this approach, one leverages machine translation to paraphrase a text while retraining the meaning. Xie *et al.* [43] used this method to augment the unlabeled text and learn a semi-supervised model on IMDB dataset with only 20 labeled examples. Their model outperformed the previous state-of-the-art model trained on 25,000 labeled examples. The BT process relies on translating the original sentence (e.g., in English) in some language (e.g., French) and then translating the resulting sentence back to the first language (English). Then, the new sentence is compared to the original sentence, if it is different, it is used as an augmented version of the original text. The BT could be run using different languages at once to generate more variations. A work by Beddiar *et al.* [2] proposed to augment six datasets for hate speech detection where back translation is shown to perform quite well and improved f1_measure and Accuracy by more than 5%.

5. *TF-IDF based word replacement*: This augmentation method was proposed by Xie *et al.* [43]. The basic idea is that words that have low TF-IDF scores are less important and thus can be replaced without affecting the ground-truth labels of the sentence. The words that replace the original word are chosen by calculating the TF-IDF scores over the whole document and assuming the words with low TF-IDF score are likely not affecting label alteration. This method has been employed with back-translation, and the model error rate was reduced from 6.5% to 4.20% for supervised learning; therefore, this method's individual effectiveness was not clear.

6. *Paraphrasing*: This is the process of rephrasing the text while keeping the same semantics. Many techniques have been used to generate paraphrases in the literature by keeping two main factors: semantic similarity and diversity of generated text [21]. For instance, [26, 9] focused on obtaining semantically similar paraphrases while [39, 40] addressed the problem of generating diverse paraphrases. Paraphrasing using BT seems to be very promising because machine-translation models were able to generate several diverse paraphrases while preserving the semantics of the text [2]. Besides, paraphrasing has been done using deep-learning models; for example, Hou et al. [11] proposed a Seq2Seq data augmentation model for the language understanding module of task-based dialogue systems. They feed the delexicalized input utterance and the specified diverse rank k (e.g., 1, 2, 3) into the Seq2Seq model as the input to generate a new utterance. Similarly, Hou et al. [10] encoded the concatenated multiple input utterances as cluster-to-cluster (C2C) by an L-layer transformer. Unlike previous Seq2Seq DA works that reconstruct utterances one by one independently, C2C jointly encodes multiple existing utterances of the same semantics and simultaneously decodes multiple unseen expressions. Both Seq2Seq and C2C have proven to show improvement in F1 scores.

7. *MixUp for Text*: Mixup is a simple yet effective image augmentation technique introduced in 2017 by Zhang *et al.* [44]. The idea is to combine two random images in a mini-batch in some proportion to generate synthetic examples for training. For images, this means combining image pixels of two different classes. It acts as a form of regularization during training. Similarly, Guo et al. [8] introduced MixUp for text, where two random sentences in a mini-batch are taken and they are zero-padded to the same length. Then, their word embeddings are combined in some proportion. The resulting word embedding is passed to the usual flow for text classification.

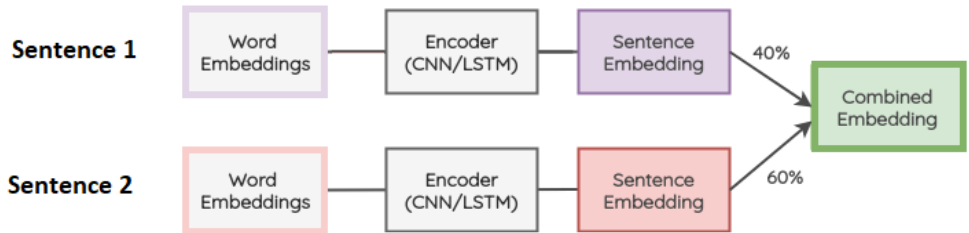

**Figure 4:** Steps of sentence MixUp using word embedding.

8. *Instance Crossover Augmentation*: This technique was introduced by Luque in his paper on sentiment analysis for TASS 2019 [29]. It is inspired by the chromosome crossover operation that happens in genetics. In general, a tweet is divided into two halves and random tweets of the same polarity (i.e. positive/negative) have their halves swapped. The hypothesis is that even though the result will be ungrammatical and semantically unsound, the new text still preserves the sentiment. This technique had no impact on the accuracy but increased the F1 score, showing that it helps minority classes such as the Neutral class with fewer tweets.

9. *Large Language Model (LLM)*: Text Generation such as GPT-3 [3] (Generative Pre-trained Transformer 3) have redefined text generation capabilities in natural language processing. In the case of GPT-3, the text generation is prompt-driven, where a user supplies a prompt to the model, and the model generates a continuation based on its learned patterns and context. The prompt serves as a starting point or context for the model to generate meaningful and contextually appropriate text. Users can experiment with prompt engineering, adjusting the wording or structure to influence the output. The process of generating text with LLMs involves sending the desired prompt to the model, which then predicts the next sequence of words based on the provided context. GPT-3, in particular, is known for its ability to generate diverse and contextually rich text across a wide range of topics. Users can fine-tune the output by modifying the prompt, adjusting parameters, or specifying desired constraints.

---

[3] https://openai.com/

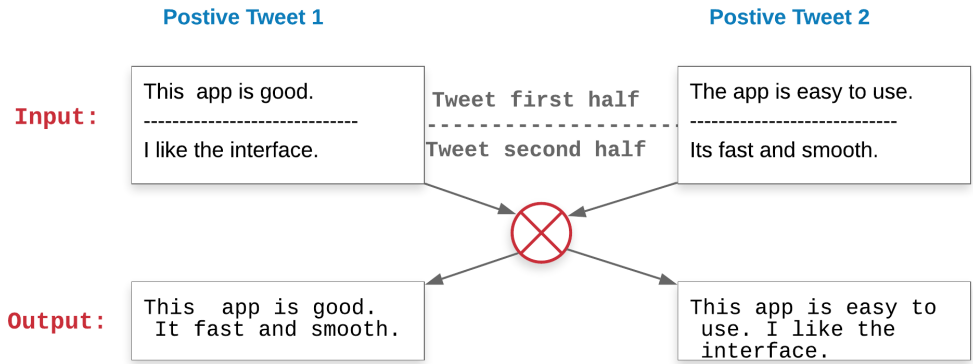

**Figure 5:** Instant crossover between two same polarity tweets.

Table 1 summarizes the aforementioned DA methods, highlighting key advantages and disadvantages. Some noise injection-based methods are enumerated below:

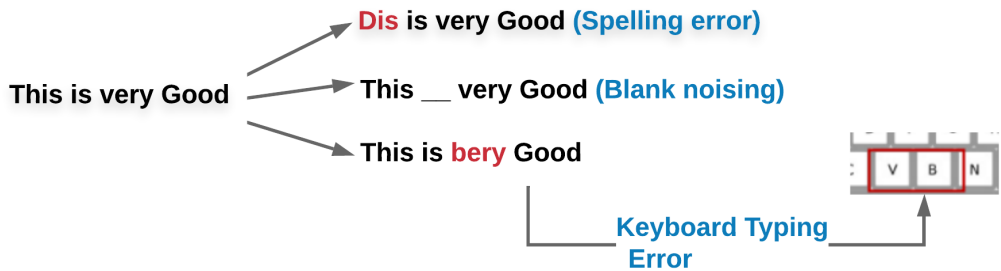

**Figure 6:** Different types of noising (e.g., spelling error, blank noising, keyboard common spelling error ).

1. Blank Noising: This method has been proposed by [43]. The idea is to replace a random word with a placeholder token i.e. "_". This is used as a way to avoid over-fitting on specific contexts as well as a smoothing mechanism for the language model. The technique helped to improve perplexity and BLEU scores (Figure 6 shows different type of noising).

2. QWERTY Keyboard Error Injection: This method simulates common errors that happen when typing on a QWERTY layout keyboard due to keys that are very near to each other. The errors are injected based on keyboard distance.

3. Unigram Noising: The idea in this method is to perform replacement with words sampled from the uni-gram frequency distribution [43]. This frequency reflects basically how many times each word occurs in the training corpus.

4. Sentence Shuffling: This is a naive technique where sentences present in a training text are shuffled to create an augmented version.

5. Random Insertion, Random Swap, Random Deletion: This consists to a random swap of any pair of words in the sentence randomly [42].

**Table 1**
Different types of NLP data augmentation practices and their general advantages and disadvantages.

| Method Name | Advantage | Disadvantage |
|---|---|---|
| WordNet | Simple to setup, fast, generates high number of augmented text [42]. | No contextual augmentation, risk of meaning alteration due to synonyms with different meanings [20]. |
| Semantic embeddings (e.g., FastText) | Simple to setup, fast, can generate high number of augmented text [42] | Risk of label alteration, no contextual augmentation [5]. The scope and part of speech of augmented words are limited. |
| BERT Mask | Diverse, contextual and higher number of augmented text [24]. | - Though the system generates a contextual synonym, however still incur label alteration and higher computational power required [2]. <br> - Require pre-trained model for training data for model finetuning [24] <br> - High training cost [24]. |
| Back-Translation, Paraphrasing | Less meaning alteration, high classifier accuracy reported [2] | - Back translation produced a maximum of one sentence per language, resulting in a small number of augmented sentences, lack of diversity, and not much control over the augmentation process [5]. <br> - The Bert machine translation requires pre-trained model [2] |
| MixUp for Text | A high number of data augmentation is possible since a large number of MixUp combinations occurs for each sentence [8]. | Meaning alteration would be increased if mixing occurred between two different polarity sentences. Moreover, it poses a similar disadvantage which uses word embedding as data augmentation[20]. |
| Random Insertion, Random Swap, Random deletion | Easy to implement [42, 24]. | High risk of meaning alteration since this method adds and deletes random words, not contextual augmentation[20]. |
| Instance Crossover Augmentation | Easy to implement, less polarity alteration since instance crossovers occur between two or more same polarity sentences [29]. | Augmented data does not contains any new words since its a crossover from original datasets, not contextual augmentation, and no classifier accuracy improvement observed, negligible improvement reported for F1 score [29]. |
| Random spelling error injection, Black noising | Easy to implement [43, 24] | This method insert noise which increase risk of meaning alteration [5]. |

## 3. Proposed Method

For performing data augmentation by replacing words in a text with other words or vector model nearest word [45, 41], prior works used synonyms as substitute words for the original words. However, synonym-based augmentation cannot produce numerous patterns from the original text. At the same time, its biggest drawback is that it might replace synonyms that could alter their meaning. Similar challenges arise when replacing the word with a synonym word produced by vector models (Word2Vec, FastText, etc.). The BERT-masked augmentation could be a solution for this since it has contextual augmentation. However, it still needs to ensure that the augmented sentences would not alter the meaning of the original sentences. Therefore, augmenting sentences by producing numerous patterns while preserving the meaning of the original sentences exists challenges.

Here, we propose contextual filtering with BERT Cosine similarity. This method can be added at the end pipeline of other methods (ex., synonym replacement, back-translation, etc.); here, we first experimented with Back-translation (BT) (Figure 7 shows the use of BT and BERT-contextual-cosine similarity, and Table 2 shows examples of augmented sentence of Figure 7 ). In this, we take an original text written in English. Then, we convert it into another language (e.g., Italian) using BERT-HuggingFace[4] translation models. Finally, we translated the Italian text back into English. After augmentation, each of these augmented sentences and corresponding original sentences is converted to contextual embedding using BERT. Finally, the Cosine similarity measures the closeness between the original and the augmented sentences.

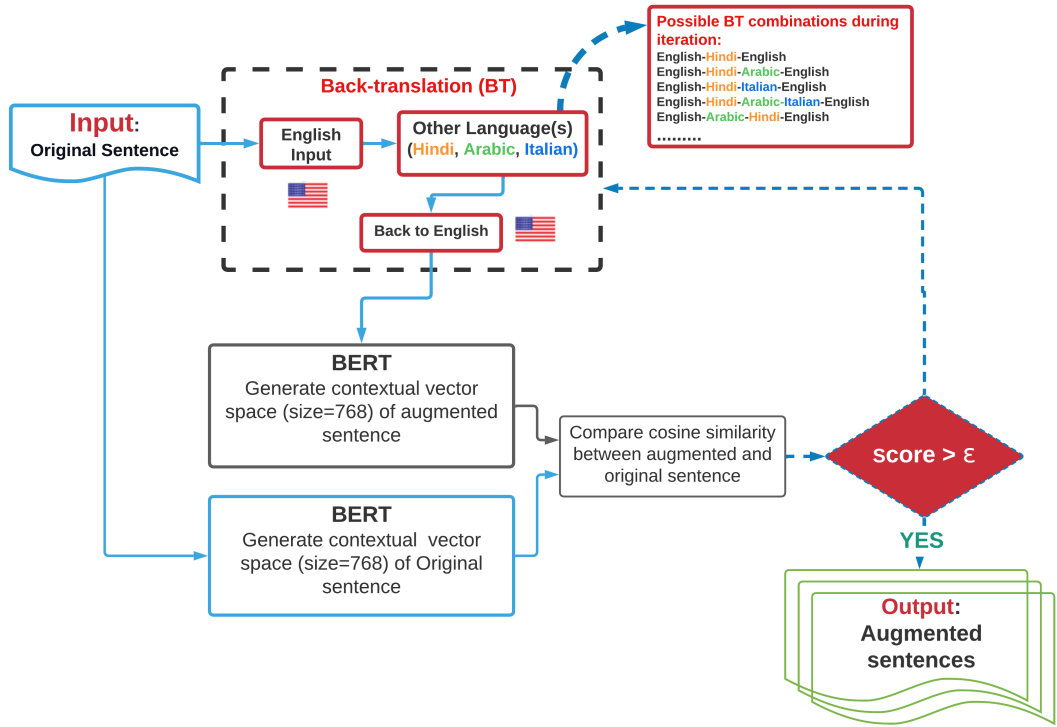

**Figure 7:** A high-level diagram of our proposed methodology. Example of a sentence expansion using Back-translation (BT). Once the new sentence is generated, BERT is used for contextual embedding. Finally, the Cosine similarity is applied to measure the closeness between the original and the augmented sentences.

**Motivating example:** We provide some samples of the original and back-translated sentences using our approach in Table 2. Examples show that augmented sentences with similarity scores of 90 or above are added to the datasets (Table 4 reference of threshold selection). For example, sentence 1 is a sample sentence from movie review datasets with a label as a positive review. After augmentation from En->Hindi->En, similarity scores show 0.84, which has not been added to the system since the original and augmented sentence no longer have the same meaning. In contrast, when the same sentence uses BT using En->Arabic->En, it produces a similarity score of 0.93, which holds the same annotation label as the original sentence; therefore has been added to the augmented dataset. Similar reflections can be seen in other datasets and examples of sentences.

---

[4] https://huggingface.co/models?search=Helsinki-NLP%2Fopus-mt

**Table 2**

Example of back-translated sentences from the used datasets, generated from source sentences with our proposed technique (Back-translation (BT), and BERT-cosine filtration ).

| Original Sentence | Original Label | BT Steps | Expanded sentences | Similarity score | Augmented sentence added to dataset |
|---|---|---|---|---|---|
| **Movie Review (MR) dataset** | | | | | |
| 1. Just an average comedic dateflick but not a waste of time | Positive | En->Hindi->En | Just an average comic story but don't waste time. | .84 | No |
| 2. Just an average comedic dateflick but not a waste of time | Positive | En->-Arabic->En | Just a regular comic story but it's not a waste of time. | .93 | Yes |
| **AskFm dataset** | | | | | |
| 3.does it make you mad that you tonsss group of hate you on myspace | Cyberbully | En->Hi->En | does it make you angry that countless grups hate you on myspace | .95 | Yes |
| 4.does it make you mad that you tonsss group of hate you on myspace | Cyberbully | En->Italian>En | Does it bother you that you have a many lively group on Myspace? | .81 | No |
| 5. You were naive | Not-cyberbully | En->Arabic>En | You was innocent | .97 | Yes |
| 6. You were naive | Not-cyberbully | En->Italian>En | you did innocent | .96 | Yes |
| 7. You were naive | Not-cyberbully | En->Arabic>En | You are stupid | .85 | No |
| 8. You and tayy should have sex and then have kids i'm not talking about now but in the future! | Cyberbully | En->Hindi->En | you and tayy have love and kids i'm not talking about now but in the future! | .84 | No |
| 9.You and tayy should have sex and then have kids i'm not talking about now but in the future! | Cyberbully | En->Italian>En | You and tayy have intercourse and then will have kids I'm talking about in the future | .93 | Yes |
| **Movie sentiment dataset (SST1)** | | | | | |
| 10.This is a film brimming with detail and nuance and one that speaks volumes about the ability of the human spirit to find solace in events that could easily crush it forever. | Positive review | En->Hindi->En | It's a movie full of detail and subtlety, and it's all about the human spirit's ability to find solace in events that could easily crush it forever. | .95 | Yes |
| **Sentence Category dataset (TREC)** | | | | | |
| 11. What fowl grabs the spotlight after the Chinese Year of the Monkey ? | Animal | En->Hidi->En | Who is the spotlight after Chinese New Year? | .88 | No |
| 12. What fowl grabs the spotlight after the Chinese Year of the Monkey ? | Animal | En->Italian->En | Which bird catches the spotlight after the Chinese year of the monkey | .92 | Yes |

---

**Algorithm 1:** Data Augmentation using proposed method: generate new sentences using BERT-back-translation + BERT-cosine similarity.

---

Process: generate new sentences using BERT-Back-translation + BERT-cosine.
**Input:** Training Dataset $D_{train} = (X_i, Y_i)$;
Create back translation model;
**for** *each sentence $(X_i, Y_i)$* **do**
    # get sentence $(X_i)$ and its label $(Y_i)$;
    # translation($X_i$): preprocess text and generate new text with translation ;
    Preprocess($X_i$);
    $X_{ibt}$ = Back_translation($X_i$) # generate back translation for $(X_i)$;
    Get BERT embedding of $(X_i)$;
    Get BERT embedding of $(X_{ibt})$;
    **if** $COS(X_i, X_{ibt}) \geq 90$ **then**
        | Add $(X_{ibt})$ as new augmented sentence;
    **end**
    # Repeat the iteration with each new language
**end**

---

**Algorithmic representation:** For instance, we refer to the text in English by $X_i$, and its translation into Italian by $S'$, while $X_{ibt}$ refers to the resulting back-translated text (i.e., in English). Back translation is therefore composed of two intermediate translations, which are given as follows (i.e., function for translation *translate()*):

$$S' = translate(model_1, X_i)$$

$$X_{ibt} = translate(model_2, S')$$

While $model_1$ is the model used for translating from English to Italian and $model_2$ is used to back translate from Italian to English. In general, back translation using one intermediate language is given by:

$$X_{ibt} = back\_translate(model_1, model_2, X_i) \tag{1}$$

Here, we have used three languages (Hindi, Arabic, and Italian) during the BT process. We keep the back-translated English text if it is different from the original English sentence (i.e. $X_i \neq X_{ibt}$). The number of iterations depends on the number of languages that have been used in the process. For example, each language can produce a maximum of one sentence and one iteration during the back-translation process of our proposed method. However, BT can provide more diversity in a chain, and changing the language order can produce different sentences. For example, 'English -> Arabic ->Italian-> English' and 'English -> Italian ->Arabic-> English' can result two different outcomes. Since we have used three languages, it gives the number of iterations based on the possible permutation of these languages (i.e. different intermediate translation models are used and therefore different results are obtained). If we express this example mathematically, we get the following:

1. First combination: 'English-> Arabic-> Italian-> English' is expressed with:

$$X_{ibt}^1 = back\_translate(model_3, model_4, model_2, X_i) \tag{2}$$

2. Second combination: 'English-> Italian-> Arabic-> English' is expressed with:

$$X_{ibt}^2 = back\_translate(model_1, model_5, model_6, X_i) \tag{3}$$

While $model_3$, $model_4$, $model_5$ and $model_6$ refer to models used for translation from English to Arabic, Arabic to Italian, Italian to Arabic and Arabic to English respectively. Note that $model_3$ and $model_6$ are different models since

---

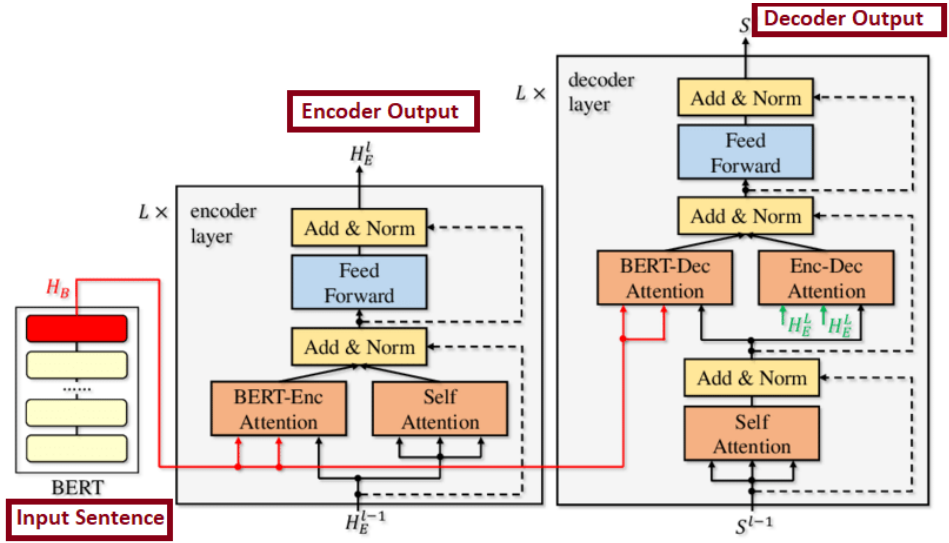

**Figure 8:** The structure of the transformer model used for machine translation tasks. We exploit this architecture for the back translation by having same layers for both the encoder and the decoder. The left and right figures represent the BERT, encoder and decoder respectively. $H_B$(red part) and $H_{LE}$(green part) denote the output of the last layer from BERT and encoder.

the source languages and the target sources are not the same (for $model_3$ source: English and target: Arabic; while for $model_6$, source: Arabic and target: English). Again, $X_{ibt}^1$ is different than $X_{ibt}^2$ but might be the same. So, each resulting translation $X_{ibt}$ is kept if $X_{ibt} \neq X_i$, with $i$ the number of possible combinations of the three languages (i.e. number of iterations).

This technique was used for neural machine translation at a large scale and yielded promising results [2]. It is based on a pre-trained transformer model (as illustrated in Fig. 8), which builds on the architecture presented in [2], where six blocks for both encoder and decoder were used. Back-translated sentences based on sampling and noised beam search were generated by translating original data into another language and back into English. Synthetic data is then concatenated to original sentences to create larger datasets.

**Cosine similarity calculation and Threshold selection:** To calculate BERT Cosine similarity, we first used BERT-base-uncased model; this has created a vector containing 768 elements. Those 768 values contain our numerical representation of a single token — which we have used as contextual word embeddings. Since each token is represented with a vector (output by each encoder), we are looking at a tensor of size 768 by the number of tokens. Finally, we have taken these tensors and transformed them to create semantic representations of the input sequence, then used our similarity metrics to calculate the respective similarity between different sequences. Cosine Distance/Similarity - It is the cosine of the angle between two vectors, which gives us the angular distance between the vectors. In general, the cosine similarity between the original sentence contextual embedding vector A and the augmented sentence contextual vector embedding B is given by:

$$sim(A, B) = cos(A, B) = \frac{A.B}{||A||.||B||} \qquad (4)$$

Finally, we can set thresholds of minimum closeness to keep relevant sentences. Before using cosine similarity, two fundamental questions arise:

1. How accurate is cosine similarity compared to Human judgment?

2. What would be the threshold limit for considering that original and augmented sentences are highly likely to present the same meaning?

**Table 3**
Pearson correlation of semantic similarity between human judgment and BERT-cosine similarity of original and augmented sentences. Back Translation is used augmentation method.

| Dataset | Human1 similarity score Vs Bert cosine similarity score | | Human2 similarity score Vs Bert cosine similarity score | |
|---------|-------------|---------|-------------|---------|
|         | Correlation | P-value | Correlation | P-value |
| MR      | 94.4        | 0.003   | 92.3        | 0.005   |
| AskFm   | 89.5        | 0.006   | 87.4        | 0.008   |
| SST1    | 88.5        | 0.005   | 87.7        | 0.007   |
| TREC    | 91.2        | 0.005   | 93.01       | 0.004   |

**Table 4**
Different thresholds for Cosine similarity and percentage of sentence annotation change between the original sentence and augmented sentence using AskFm datasets.

| Method Name | Cos. value >.50 | Cos. value >.70 | Cos. value >.90 |
|-------------|-----------------|-----------------|-----------------|
| Wordnet | 4.0% | 2.0% | 0.1% |
| FastText | 2.0% | 2.5% | 0.3% |
| BERT Mask | 6.0% | 3.0% | 0.5% |
| Back-Translation | 1.5% | 0.5% | 0.3% |

To address the first question, we conducted a comparison between the human similarity score and the BERT-cosine score of both original and augmented sentences. Specifically, we utilized 100 original sentences and 100 augmented sentences generated by BT augmentation methods from each dataset, including MR, TREC, SST1, and AskFM. Two human judges with expertise in this area manually evaluate and provide similarity scores for each original and augmented sentence separately while keeping their evaluations concealed from each other. The similarity scores were provided in the range of 0 to 1, where 0 indicated no similarity between the two sentences, and 1 indicated the highest possible similarity. Finally, we computed the Pearson coefficient. The correlation and P-value of semantic similarity between human judgment and BERT-cosine similarity of both original and augmented sentences are displayed in Table 3. Our findings indicate that both Human1 and Human2 exhibited higher correlation scores and smaller P values, indicating the statistical significance of the results. In other words, the application of BERT-cosine similarity can potentially identify similar sentences with a level of accuracy comparable to that of human judgment.

To answer question 2, we have performed empirical analysis through trials, and incremental error-correcting procedures have been employed to set up threshold values by manually checking the quality of the first 500 posts after each incremental choice of the threshold.

Table. 4 shows that the above 0.90 thresholds produced fewer label alterations during augmentation. Therefore, in our experiment, we selected 0.90 as a minimum threshold for similarity. So, augmented sentence is kept only if:

$$(sim(X_i, X_{ibt}) = cos(X_i, X_{ibt})) > 0.9 \tag{5}$$

Furthermore, we assessed the effectiveness of using a threshold selection of 0.90 for the BERT-cosine similarity metric in evaluating the semantic similarity between two sentences. To accomplish this, we utilized the STSS-131 dataset, which comprises ground truth data from the study conducted by O'shea et al. titled 'A new benchmark dataset with production methodology for short text semantic similarity algorithms'. Specifically, we employed the Pearson correlation coefficient to evaluate the correlation between the BERT-cosine similarity scores and the human judgments provided in the dataset. For instance, given two sentences S1 and S2, we compared the human assessment of the semantic similarity between them in the STSS-131 dataset with the BERT-cosine similarity score. The results of the Pearson correlation and P-value scores, before and after using the 0.90 threshold filtration, are presented in Table 5.

Initially, we compared the BERT-cosine similarity scores with human judgment for S1 and S2 and found that the Pearson correlation and P-value scores were 0.867 and 0.003, respectively, indicating a high correlation. These findings are consistent with earlier results reported in Table 3. We then generated an augmented sentence for each of the STSS-131 dataset sentence pairs and calculated the BERT-cosine similarity score, Pearson correlation, and P-value scores. Subsequently, we filtered out the sentence pairs with a cosine similarity value of less than 0.90 and repeated the calculation. The ultimate outcomes of the experiment demonstrated a statistically significant improvement in the

**Table 5**
STSS-131 dataset Pearson correlation of semantic similarity between human judgment and BERT-cosine similarity.

|  | Pearson correlation of semantic similarity between human judgment and BERT-cosine similarity. | |
|---|---|---|
|  | Correlation | P-value |
| STSS-131 dataset without augmentation | 0.867 | 0.003 |
| STSS-131 dataset after augmentation | 0.883 | 0.002 |
| STSS-131 dataset after augmentation and Cos. value > 0.90 | 0.89 | 0.002 |

Pearson correlation after applying the cosine filtration. Specifically, the Pearson correlation scores before and after cosine filtration were 0.88 and 0.89, respectively. These results demonstrate that the 0.90 threshold leads to fewer label alterations and improves the accuracy of the BERT-cosine similarity metric.

In Section 4.2, we provide additional details about the experiment and the manual checking scores for label alteration using other datasets. Specifically, 12 and 11 present the results of the label alteration experiments and the average cosine similarity scores before and after filtering, respectively. These results demonstrate that the 0.90 threshold leads to fewer label alterations and improves the accuracy of the BERT-cosine similarity metric.

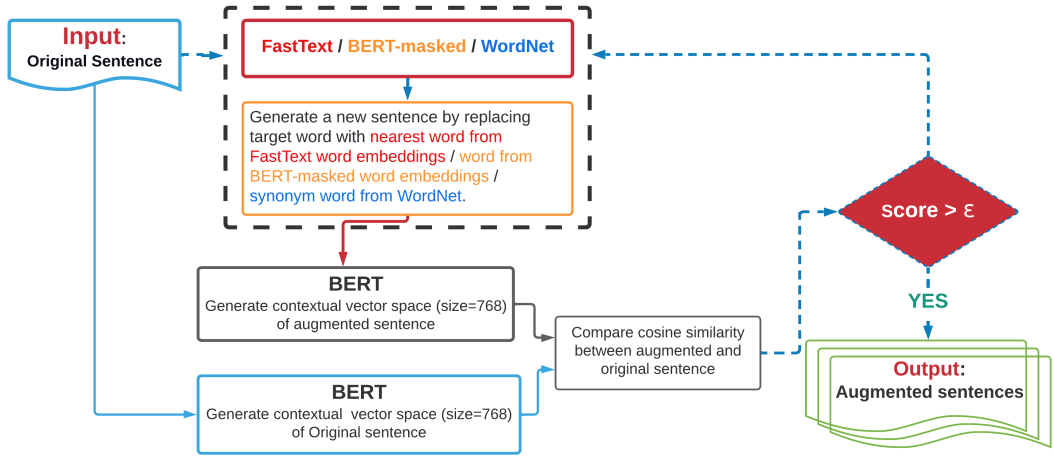

**Figure 9:** Example of a sentence expansion using synonym replacement using other practices (FastText, WordNet, BERT-masked). Once the new sentence is generated, BERT is used for contextual embedding. Finally, the Cosine similarity is applied to measure the closeness between the original and the augmented sentences.

## 3.1. Alternative Experiment

Besides back-translation, we have tested our proposed method using other word replacement practices, FastText, WordNet, and BERT-contextual augmentation (Figure 9). This section presents a brief description of Algorithm 2 for text generation process using FastText + BERT-cosine, WordNet + BERT-cosine, and BERT-mask + BERT-cosine. Here the train dataset $D_{train}$ contains M number of sentences and their corresponding labels $(X_i, Y_i)$. Each sentence may have N number of words $(X_i W_N)$.

For FastText and WordNet, our algorithm targets each word at a time $(X_i W_n)$ and its synonym and replaces the targeted word with the synonyms, which creates augmented sentence $X_{i'}$ as follows:

$$X_{i'} = replace(X_i, X_i W_n, synonym(X_i W_n)) \qquad (6)$$

After that, BERT embedding is generated for the original and augmented sentence. If the contextual similarity is greater than or equal to 90% (i.e. $cos(X_i, X_{i'}) > 0.9$), then the algorithm adds that sentence as an augmented sentence. The new augmented sentence has the same label as the original sentence $(X_{i'}, Y_i)$. This process repeats for each word (we have considered the top 5 nearest words for FastText).

For BERT-mask, first, we made 15% sample tokens as mask tokens, which was replaced by the predicted tokens. For a sentence $X_i$ with $N$ tokens, $N * 15\%$ of mask tokens are predicted and replaced as follows, while Nm is the number of mask tokens:

$$Nm = N * 0.15 \qquad (7)$$

In the MLM task, the goal is to predict the original token Wt using contextual information from other tokens. As illustrated by algorithm 2, for a sentence $X_i$ with $N$ words, the token $W_t$ is to be replaced by a mask token and the token representation for sentence $X_i$ would be $X_i[W_1, W_2, W_{t[mask]}, ... W_n]$. The predicted probability Pp for the token $Wt$, given a set of candidate tokens $Wp$, can be calculated as follows:

$$Pp = \text{softmax}(W_t | W_p) \qquad (8)$$

This 15% of making done scholastically and the predicted token could be different for each iteration; therefore, we have used 50 iterations for maximizing the diversity of augmented sentences. Finally, we applied BERT-cosine similarity extent for each iteration and constructed augmented sentences.

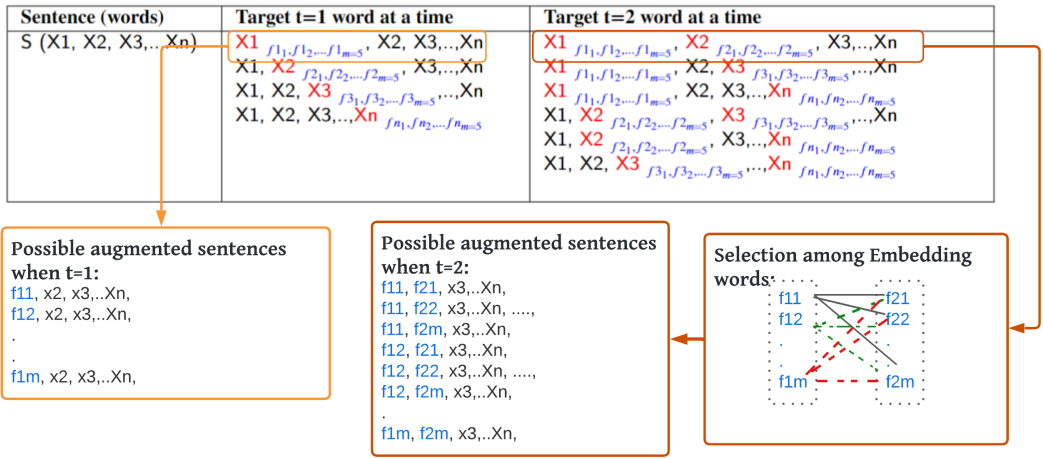

**Figure 10:** Selection of target words for synonym replacement and possible augmented sentences using FastText. Here, 'S' represents the sentence, and $X_n$ represents the number of words. The red words are the target words for synonym replacement, and $fn_m$ is the list of nearest words of $X_n$. Here we have limited the top 5 closest words from word embeddings for synonym replacement.

Table 6 shows the sentence augmentation using FastText word embeddings and BERT-cosine similarity filtration. From the example, we see that several embedding words have been used that are not suitable in context to the original sentence; for example, gay synonyms used as (brave, jolly, and festive) are not suitable. Therefore, after using BERT-cosine filtration, those sentences have been successfully removed. In addition, Table 7 also shows promising results using another alternative practice (BERT MLM word replacement) and BERT-cosine filtration from MR, AskFm, SST1, and TREC datasets.

---

**Algorithm 2:** Alternative experiment algorithm: Wordnet + BERT-cosine, FastText + BERT-cosine, and BERT-MLM + BERT-cosine

---

Input:Training Dataset $D_{train}$;
\# Process 1: generate new sentence using FastText + BERT-cosine;
Read dataset line by line;
**for** *each sentence $(X_i, Y_i)$* **do**
    **for** *each word: $X_i W_n$* **do**
        Find $j$ nearest synonyms for $X_i W_n$ denoted by $W_{nj}$;
        **for** *j in range(5):* **do**
            New sentence($X_{i'}$): replace $X_i W_n$ with its synonym $W_{nj}$ ;
            Get BERT embedding of ($X_i$);
            Get BERT embedding of ($X_{i'}$);
            **if** $COS(X_i, X_{i'}) \geq 0.90$ **then**
                | Add ($X_{i'}$) as new augmented sentence;
            **end**
        **end**
    **end**
**end**
\# Process 2: generate new sentence using WordNet+BERT-cosine ;
Read dataset line by line;
**for** *each sentence $(X_i, Y_i)$* **do**
    **for** *each word: $X_i W_n$* **do**
        Find j nearest synonyms for $X_i W_n$ denoted by $W_{nj}$;
        New sentence($X_{i'}$): replace $X_i W_n$ with its synonym $W_{nj}$ ;
        Get BERT embedding of ($X_i$);
        Get BERT embedding of ($X_{i'}$);
        **if** $COS(X_i, X_{i'}) \geq 90$ **then**
            | Add ($X_{i'}$) as new augmented sentence;
        **end**
        Repeat the process with last synonym $W_{nj}$;
    **end**
**end**
\# Process 3: generate new sentence using BERT-MLM + BERT-cosine ;
Read dataset line by line ;
**for** *each sentence $(X_i, Y_i)$* **do**
    15% of tokens are sampled to be predicted and replaced with [MASK] token: $X_i[W_1, W_2...W_{[mask]}....W_n]$;
    New sentence($X_{i'}$): replace [mask] token ($X_i W_{[mask]}$) with predicted tokens $W_p$;
    Get BERT embedding of ($X_i$);
    Get BERT embedding of ($X_{i'}$);
    **if** $COS(X_i, X_{i'}) \geq 90$ **then**
        | Add ($X_{i'}$) as new augmented sentence;
    **end**
    **else** Repeat the process for 50 iterations;
**end**

---

## 3.2. GPT3 Text Augmentation:

In the augmentation process leveraging the GPT3 API for hate speech data, a Python script is used to harness the language generation capabilities of the GPT3 model. Beginning with the set-up of the GPT3 API access and the installation of the OpenAI Python package, the function, named generate-gpt3-augmentation, facilitates the interaction with the GPT-3 engine. The function accepts a prompt as input, with a structured template designed to solicit augmented

---

**Table 6**

An example of sentence augmentation using alternative practice(FastText word replacement), and BERT-cosine filtration from hate dataset.

| Original Sentence | Target words (Nearest FastText embeddings words) | Expanded sentences | Similarity score | After BERT-cosine filtration |
|---|---|---|---|---|
| **Targeting one word at a time** | | | | |
| He is gay | gay, (lesbian, queer, brave, festal, sunny) | he is lesbian | 0.95 | he is lesbian |
| | | he is queer | 0.93 | he is queer |
| | | he is brave | 0.70 | |
| | | he is festal | 0.65 | |
| | | he is sunny | 0.74 | |
| **Targeting two word at a time** | | | | |
| He is gay | he (she, afterwards, etc.); gay (lesbian, homosexual, brave, jolly, etc.) | she is lesbian | 0.93 | she is lesbian |
| | | she is homosexual | 0.91 | she is homosexual |
| | | she is brave | 0.78 | |
| | | she is jolly | 0.63 | |
| | | afterwards is lesbian | 0.77 | |
| | | afterwards is homosexual | 0.69 | |
| | | afterwards is sunny | 0.55 | |
| | | afterwards is jolly | 0.54 | |

**Table 7**

Example of sentence augmentation using alternative practice (BERT MLM word replacement), and BERT-cosine filtration from MR, AskFm, SST1, and TREC datasets.

| Original Sentence | Original Label | Expanded sentences | Similarity score | Augmented sentence added to dataset |
|---|---|---|---|---|
| **Movie Review (MR) dataset** | | | | |
| 1. Just an average comedic dateflick but not a waste of time | Positive | just an average comedic dateflick, a rush of time. | 0.85 | No |
| 2. Just an average comedic dateflick but not a waste of time | Positive | basically an average comedic relief now not a waste of time. | 0.95 | Yes |
| **AskFm dataset** | | | | |
| 5. You were naive | Not-cyberbully | you were him | 0.81 | No |
| 6. You were naive | Not-cyberbully | you call naive | 0.94 | Yes |
| 8. You and tayy should have sex and then have kids i'm not talking about now but in the future! | Cyberbully | now sure tayy better have sex have kids im not talking normal now but in the future! | .94 | Yes |
| **Movie sentiment dataset (SST1)** | | | | |
| 10. This is a film brimming with detail and nuance and one that speaks volumes about the ability of the human spirit to find solace in events that could easily crush it forever. | Positive review | this promises a film brimming full detail and nuance and one that holds volumes the ability or the human spirit the find peace in events can could easily corrupt men forever | 0.95 | Yes |
| **Sentence Category dataset (TREC)** | | | | |
| 11. What fowl grabs the spotlight after the Chinese Year of the Monkey ? | Animal | 'what fowl growled at claw across the chinese year of the monkey?' | 0.92 | Yes |
| 12. What fowl grabs the spotlight after the Chinese Year of the Monkey ? | Animal | such fowl missed the spotlight after year of the moon? | 0.87 | No |

versions of a given sentence, prefacing the original sentence with contextual information. For instance, the prompt might read, "Given the sentence 'The tweet contains hate speech: YourOriginalSentence,' provide 20 augmented versions that may contain hate speech or not." Example of a sample prompt for generating augmented sentences for hate /non-hate:

```
model_id: 'gpt-3.5-turbo-0613'
temperature: 0
```

**Table 8**

Example of sentence augmentation using GPT3 from AskFm datasets.

| Original Sentence | Augmented Sentence | Label |
|---|---|---|
| sorry im from LA | I apologize; I'm from the city of Los Angeles | 0 |
| Is your schlong bigger? | Do you have a bigger dick? | 1 |
| You judgmental hoe | You bigot whore | 1 |
| What can I ask about then | What other topics are open for discussion? | 0 |

```
max_tokens: 256
llm_mode: 'chat'

# chat mode: system_role, user_role and assistant_role.
# instruction mode: instruction, input and response.
prompts:
  - system_role: "As a data moderator, your role is to augment new sentences based on provided
      user-generated content.
  **

  - if the content has label 0 that means it does not contain any
      hate/cyberbulling/improper/offensove content. Now genrate 20 sentences for each sentences,
      makesure augmented data is align with the corresponding label.

- if the content has label 1 that means it does contain any hate/cyberbulling/improper/offensove
      content. Now genrate 20 sentences for each sentences, makesure augmented data is align with
      the corresponding label.
  "
  - user_role: "The input text is: "
```

The API call is then initiated using this prompt, specifying parameters such as the choice of the text-davinci-003 engine, a max-tokens limit to manage text length (1000), and a temperature of 0 to favor more deterministic output. The resulting response from the API call is processed to extract the augmented sentences, which are subsequently post-processed to ensure cleanliness and relevance. Finally, the augmented sentences are ready for further analysis, providing a diversified set of potential inputs for hate speech detection models. Table 8 illustrates an instance of sentence augmentation achieved through GPT-3 using AskFm datasets.

## 4. Experiment METHODOLOGY

We chose five benchmark datasets for testing our proposed augmentation method. The overall experimentation methodology includes a four-stage process: (i) dataset selection, (ii) data augmentation, (iii) comparison between different augmentation methods, and (iv) classification performance comparison (accuracy, F1 score), and error analysis.

1. First, we select five publicly available benchmark datasets to test data augmentation using different methods as well as the classifier accuracy for augmented and non-augmented datasets.

2. Second, we expand the selected datasets with different existing augmentation methods (e.g., synonym replacement with WordNet and FastText , Back-translation, BERT mask) and apply our proposed methods (e.g., Back-translation + BERT cosine). For the testing phase with Large Language Models (LLMs), specifically GPT-3, we utilize the GPT-3 API interface to generate augmented sentences.

3. After data augmentation, we compare our proposed method with other popular methods in terms of: the number of newly generated data; and the quality of the expanded sentence (how close is it to the original sentence).

4. Next, we experiment ML performance with all expanded datasets and not expanded datasets for our proposed method and other existing methods. For that, we use: Accuracy, F1_measure. We exploit different classifiers such as NB, LR, CNN model, and BERT-base. We contrast the results and evaluate the data augmentation process for all four datasets.

**Table 9**
Some important information on the used datasets with some samples from original dataset

| Dataset | Size | Classes | Number of samples per class | Examples |
|---|---|---|---|---|
| **AskFm** | 9998 | 2 classes | Not cyberbullying: 8789 *Cyberbullying: 1209 | 'You are bautiful' *'no india can code an entire software and whatever he codes is buggy and if he does the fonrt end that will be like 4th world shit that is india ha-haha' *'why the sand niggers should vote for obama' |
| **HASOC-2021** | 3843 | 2 levels of classes | Not offensive: 1342 *Offensive: 2056 | 'i mean it worked for gun control right url' *'user oh noes tough shit b..ch' *'obama wanted liberals amp illegals to move into red states as f..k' |
| **SST1** | 10605 | 5 classes | very negative, negative, neutral, positive, very positive | Very Positive: 'One of the greatest family-oriented, fantasy-adventure movies ever'. Negative: 'Emerges as something rare, an issue movie that 's so honest and keenly observed that it doesn't feel like one'. |
| **MR** | 2000 | 2 class | Positive review: 1000, *Negative review: 1000 | If you sometimes like to go to the movies to have fun, wasabi is a good place to start. *The story is also as unoriginal as they come, already having been recycled more times than I'd care to count.' |
| **TREC** | 5452 | 6 question class | Abbreviation, Description, Entities, Human beings, Locations, Numeric values | Human: manner How did serfdom develop in and then leave Russia? Entities: cremat What films featured the character Popeye Doyle ? Abbreviation: exp What is the full form of .com ? |

Finally, we perform error analysis for both expanded and not-expanded datasets. Besides, we consider semantic fidelity to evaluate the use of BERT-cosine filtration. We measure the statistical significance of BERT-cosine similarity scores compared to human judgment.

## 4.1. Datasets

*HASOC 2021 hate speech dataset:* is an English Twitter dataset for hate and offensive language. The HASOC [30] task organizer already annotated datasets for English subtask A. For instance, if the Twitter post contains any hate or offensive word or represents any offensive context, it is considered HOF (hate or/and offensive), otherwise NOT. The total size of the dataset is 3843, among which 2051(65%) contain HOF and the rest 1342 (35%) NOT.

*AskFm dataset:* contains data from Ask.fm website [5] which serves for asking questions and getting answers either anonymously or non-anonymously. This dataset was employed to detect cyberbullying in question-answer pairs and is composed of 10000 entries, where label 1 is assigned if the entry corresponds to cyberbullying and 0 otherwise [14].

*SST1:* is the Stanford Sentiment Treebank movie review dataset with five categories labels (very negative, negative, neutral, positive, very positive, respectively) [38]. The dataset contains 10,605 processed snippets from the original pool of Rotten Tomatoes HTML files.

*MR* : is a movie review dataset for detecting positive/negative reviews [35]. The total dataset size is 2000, which contains a balanced number of positive and negative reviews. This dataset is based on user ratings from rotten tomatoes where 'Rotten' indicate negative, and 'FRESH' represents positive review.

*TREC* : is a question dataset to categorize a question into six question types (ABBR - Abbreviation, DESC - Description and abstract concepts, ENTY - Entities, HUM - Human beings, LOC - Locations, NYM - Numeric values) [25]. The dataset contains 5452 labeled questions for train and 500 for test samples.

### 4.1.1. Dataset Preprocessing

Our cyberbullying and hate speech datasets were pre-processed using standard NLP tools, mainly removing unidentified characters, symbols, and tab tokens (e.g., @,0-9, #,+ etc.) and converting all characters to lower case. Many abbreviated words and short forms of social network slang are replaced with their original terms (e.g., u = you, em = them, tbh = to be honest, etc.). Besides, stop-words, which are generally the most common words in a language, are removed. We have used NLTK's[6] list of English stop-words as reference. However, we have modified those stop words since some of them are essential to determine cyberbullying and hate-speech. For example, in some cases, person indicator words (e.g., he, she, his, her, you, yourself, etc.) could be considered stop-words; however, those play a crucial role in cyberbullying detection.

We have not pre-processed review datasets (SST1 and MR) and sentence categorical datasets (TRACK) since they were already pre-processed and much cleaner.

## 4.2. Expansion Experiment Using Different Methods

In this section, we present our experiments with four popular data augmentation methods. For each of these methods, we escalate the possible limitations and use BERT cosine similarity to keep only those augmented sentences that are contextually 90% close to the original sentences.

After expansion using different existing methods, we observed two essential insights. First, every technique is quite successful in generating a large number of augmented sentences. From Table. 10, we can see that FastText nearest word replacement produced 99 times augmentation, WordNet synonym replacement 78 times, BERT Mask contextual augmentation 43 times, and Back-Translation made 4.9 times compared to the original dataset. For BERT-mask, we have only allowed the top 50 nearest word replacements; therefore, it has not augmented as large compared to WordNet and FastText. On the other hand, Back-Translation based augmentation showed the lowest number of augmented sentences because back-translation only generates one sentence per language iteration. Here, we have used five languages for this back translation: English, French, Portuguese, Italian, and Polish. If we consider more languages, then the augmentation size would be larger.

Second, it is worth knowing which augmentation is more reliable for supervised learning and which does not alter annotation labeling during augmentation. Since augmentation is a process without any human intervention, there would be a chance of label alteration. If that happens, there might be thousands of incorrectly labeled data in the trained sample. Hence, we manually check the first 500 sentences for each augmentation method.

Table. 12 shows the results of labeled alteration using different augmentation methods for all four datasets. We notice that all four augmentation methods produced many annotation alterations, which likely would cause much error in the machine learning performance. Among all these methods, we found that label alteration happens mostly for hate/cyberbullying data compared to non-hate sentences. We show examples of label alteration for each augmentation method in Table. 13. We see that the original sentence "Wait gay, why didn't you tell him that?' was labeled as hate/cyberbullying; however, the Back-Translation augmentation produced a non-cyberbullying sentence that is: 'Wait, why didn't you tell him that?'. Since the original sentence was labeled hate, the augmented sentence was also labeled as hate, though it should be labeled as non-hate. By manual inspection of each method, we have found the label alteration rate for WordNet is 5%, for fasText is 3%, for BERT-mask is 5%, and for Back-Translation is 1.5% for the hate/cyberbullying section. After BERT contextual cosine similarity was applied to each method, the result was promising, and labeled alteration was reduced to 0.4%. However, GPT-3, without any modification, exhibits an almost negligible label alteration in augmented sentences.

---

[5] https://ask.fm/
[6] https://gist.github.com/sebleier/554280(accessed Dec 30, 2020)

**Table 10**
Size of datasets before and after expansion using different methods: WordNet, FastText, Back-Translation, BERT mask, and WordNet + FastText + BERT mask + BERT Cosine Similarity (BT refers to Back-Translation and Cos.Sim to the cosine similarity).

| | AskFM | HASOC | SST1 | TREC | MR | Avg. Expansion size |
|---|---|---|---|---|---|---|
| **Before expansion** | 10k | 3.8k | 10.6k | 5.4k | 2K | - |
| **After WordNet** | 623K | 338k | 943k | 480k | 143.3k | 89 times |
| **After FastText** | 693K | 376K | 1049k | 534k | 159K | 99 times |
| **After BT** | 20.3K | 10.64K | 30.7k | 15k | 4.7K | 2.9 times |
| **After BERT mask** | 301K | 159K | 455k | 232 k | 69.8K | 43 times |
| **After WordNet + BERT Cos.Sim** | 497K | 265K | 752 | 380k | 115K | 71 times |
| **After FastText + BERT Cos.Sim** | 336K | 178K | 508k | 249k | 78.6K | 48 times |
| **After BERT mask + BERT Cos.Sim** | 70K | 36K | 100.6k | 50.3k | 17K | 10 times |
| **After BT + BERT Cos.Sim** | 9.1K | 4.9K | 13.78 | 6k | 2.2K | 1.3 times |
| **After WordNet+ FastText+ BT + BERT mask + BERT Cos.Sim** | 850K | 323K | 870k | 500k | 213.1K | 100 times |
| **GPT3** | 200K | 76K | 212k | 108k | 40K | 20 times |

**Table 11**
Average cosine similarity between original sentences and augmented sentences using different methods.

| **Method Name** | **AskFm** Avg. Cos | **HASOC** Avg. Cos. | **SST1** Avg. Cos. | **MR** Avg. Cos | **TREC** Avg. Cos |
|---|---|---|---|---|---|
| WordNet | .94 | .93 | .93 | .94 | .94 |
| FastText | .87 | .89 | .87 | .88 | .88 |
| BERT Mask | .80 | .81 | .79 | .80 | .80 |
| Back-Translation | .94 | .93 | .93 | .92 | .94 |
| After WordNet+ FastText+ BT + BERT mask + BERT Cosine Similarity | .96 | .98 | .96 | .96 | .97 |
| After GPT3 | .95 | .97 | .95 | .97 | .98 |

**Table 12**
Percentage of change of sentence annotation between the original and the augmented sentences (Human checking). BT refers to Back-Translation, Pos. refers to a positive review and Neg. refers to a negative review.

| **Method Name** | **AskFm** | | **HASOC** | | **SST1** | | **MR** | |
|---|---|---|---|---|---|---|---|---|
| | Hate | Non-hate | Hate | Non-hate | Pos. | Neg. | Pos. | Neg. |
| WordNnet | 5.0% | 1.0% | 4.0% | 1.0% | 5.0% | 1.0% | 4.0% | 0.0% |
| FastText | 3.0% | 1.0% | 3.2% | 0.0% | 3.5% | 0.5% | 3.0% | 1.0% |
| BERT Mask | 5.0% | 1.5% | 5.0% | 1.0% | 5.0% | 1.0% | 5.0% | 2.0% |
| Back-Translation | 1.5% | 0.0% | 1.0% | 1.0% | 2.0% | 0.5% | 1.0% | 0.0% |
| After WordNet+ FastText+ BT + BERT mask + BERT Cosine Similarity >0.90 | 0.4% | 0.0% | 0.2% | 0.1% | 0.6% | 0.0% | 0.4% | 0.0% |
| After GPT3 | 0.0% | 0.0% | 0.2% | 0.0% | 0.0% | 0.0% | 0.0% | 0.0% |

**Table 13**
Label alteration examples from the AskFm datasets (BT refers to back-translation).

| Method | Original Sentence | Label | Augmented Sentence | Label |
|--------|------------------|-------|--------------------|-------|
| BERT Mask | does it make you mad that you tonss group of hate you on myspace | 1 | does it make you mad that you tonss group of love you on myspace | 1 |
| WordNet | michelle i am your obama so now we s*x | 1 | michelle i am your obama so now we love | 1 |
| FastText | you can shove that up your rude as* dumb as* bi**h | 1 | yoon-farm can shove that on-farmp yoon-farmr ron-farmde as* don-farmmb donkey | 1 |
| Back-Translation | Wait gay, why didn't you tell him that? | 1 | Wait, why didn't you tell him that? | 1 |

## 4.3. Classification Experiment

To classify each dataset for identifying hate/non-hate speech, sentiment (positive/negative reviews), and sentence categorization, we evaluated primarily deep learning classifiers CNN and BERT. We excluded logistic regression and other non-deep learning models due to their inferior performance compared to deep learning approaches. We followed basic NLP procedures during this classification experiment and utilized feature engineering techniques, specifically word embedding, for the CNN model.

### 4.3.1. Classification Architecture

Once our data was pre-processed, we performed the binary classification. Initially, we employed a random split of the original dataset into 70% for training, 10% for validation, and 20% for testing. All the results in this study have followed the same test setup. A similar procedure was followed for the HASOC, SST1, TREC, and MR datasets as well. This was very important because if the test data varied from one method to the next, that would be a significant flaw in the methodology. Two types of classifiers were implemented: Convolution Neural Network (CNN), and BERT.

We followed the convolutional neural network (CNN) design as outlined in Kim (2014) [18], wherein the initial layer integrates the words constituting the post (limited to 70 words) through concatenation. However, a key modification was implemented: each word is encoded using its FastText embedding, resulting in a 300-dimensional vector representation. The concept of Word Embeddings Features involves mapping individual tokens to vectors of real numbers. This mapping is designed to quantitatively and categorically encapsulate the semantic similarities among linguistic terms, leveraging their distributional characteristics across extensive corpora through machine learning (ML) or similar dimensionality reduction methods. In this study, we utilized the FastText pre-trained word embeddings [7]. For the convolutional aspect of our model, we applied a 1D convolution operation with a kernel size of 3, followed by a max-over-time pooling operation across the feature map, and integrated a dense layer comprising 50 units. To regularize, we introduced a dropout mechanism before the final layer, along with an l2-norm constraint on the weight vectors. The structure of our CNN model is depicted in the referenced figure.

For the BERT text classifier, we used "bert-base-uncased" model. The BERT model has 12 layers, 768 hidden states, and 12 heads. We used the pooled representation of the hidden state of the first special token([CLS]) as the sentence representation. To mitigate overfitting, a dropout probability of 0.1 is applied to the sentence representation before it is passed to the Softmax layer for classification. Further refining the model's training and evaluation process, we experimentally selected a set of hyper-parameters for best performance of the model. The learning process was guided by a fixed learning rate of 1e-5, with the learning rate scheduling set to constant, ensuring a steady optimization path. To accommodate computational constraints, the maximum sequence length was limited to 256 tokens. The model training was designed to be relatively swift, set to complete within a single epoch, to prevent overfitting. Consistency and reproducibility were key concerns, addressed by setting a random state value of 102 for both model initialization and train-test splitting. Training, validation and prediction batches were processed with a size of 32, carefully balancing speed and memory efficiency to optimize the overall training workflow.

The implementation details are reported on the GitHub page [8].

---

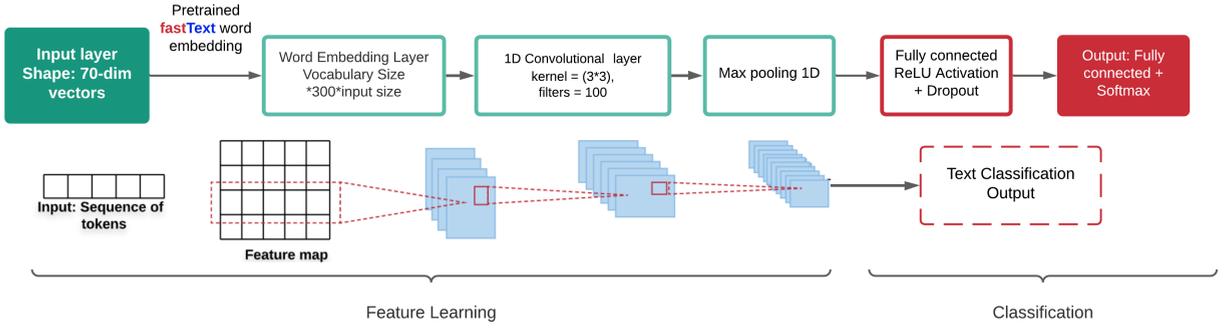

**Figure 11:** The architecture of our proposed text classification using CNN and FastText.

### 4.3.2. Performance metrics

To demonstrate the performance of our proposal, we calculate the accuracy and F_Measure as follows:

*F_Measure*: determines the harmonic mean of precision and recall by giving information about the test's accuracy. It is expressed mathematically as:

$$F\_Measure = 2 * \frac{Precision * Recall}{Precision + Recall} \qquad (9)$$

*Accuracy*: measures the percentage of correct predictions relative to the total number of samples. It can be expressed as:

$$Accuracy = \frac{\text{Correct Predictions}}{\text{Total Predictions}} = \frac{TP + FN}{TP + FN + TN + FP} \qquad (10)$$

Where TP, FN, TN, and FP correspond to true positive, false negative, true negative, and false positive, respectively.

### 4.3.3. Classification Results

The experiment was conducted in three stages. Initially, we evaluated a dataset without any augmentation. Subsequently, we applied four distinct augmentation techniques: WordNet, FastText, Back-translation, and BERT masking. In the final stage, we applied BERT-base cosine similarity filtering to each augmented dataset. Additionally, we assessed the augmentation performance of GPT-3.

Table 14 shows results of not using the data augmentation. In this analysis, we chose the optimal classifier architecture. The comparison reveals that, in the absence of data augmentation, BERT marginally surpasses CNN, showing a slight improvement of 0.9% in accuracy and 1.0% in F1 score. Consequently, for experiments involving data augmentation, our focus has been solely on BERT, the state-of-the-art (SOTA) model demonstrating the best performance.

Table 15 exhibits the experiment results of non-augmented and augmented datasets with and without BERT-cosine for four different datasets: AskFm, HASOC, TREC, and SST1. The examination of BERT-based classifier performance across these four distinct datasets—AskFm, HASOC, TREC, and SST1—reveals insights into the effects of data augmentation and the role of BERT-cosine similarity filtration. Initially, the impact of data augmentation methods such as WordNet, FastText, and BERT, when applied without BERT-cosine similarity, shows minimal to no improvements in accuracy and F1 scores. For example, BERT-mask data augmentation for AskFm has resulted in a drop of 0.3% in accuracy and 0.2% in F1. This suggests that the mere addition of augmented sentences does not guarantee enhanced model performance, emphasizing the necessity for quality over quantity in augmentation strategies.

Among the various data augmentation methods evaluated, back-translation demonstrated the best performance, showing a 0.5% improvement despite generating a smaller number of augmented sentences compared to WordNet, FastText, and BERT-masking methods. One possible explanation for this is that, although back-translation produces fewer sentences, it results in less label alteration compared to other methods, as observed in the previous table 12. This minimal label alteration may contribute to its superior performance. In contrast, BERT-masking methods, which showed the highest rate of label alteration, exhibited the lowest improvement in classifier accuracy.

**Table 14**

Classifier Accuracy (%) and F1_Measure (%) for AskFm (class: cyberbullying, not-cyberbullying) and MR (class: positive and negative review) not expanded datasets using different classifiers.

| Classifier and Feature Name | Not Expanded AskFm | | Not Expanded FormSpring | |
|---|---|---|---|---|
| | Acc. | F1 | Acc. | F1 |
| CNN + Word Embedding | 90.2 | 89 | 96.2 | 95.8 |
| BERT-base-uncased | **91.1** | **90** | **96.7** | **96.4** |

**Table 15**

Classifier Accuracy (%) and F1 scores (%) of BERT-base classification for original and expanded datasets.

| Dataset name | AskFm | | HASOC | | SST1 | | TREC | |
|---|---|---|---|---|---|---|---|---|
| | Acc. | F1 score | Acc. | F1 | Acc. | F1 | Acc. | F1 |
| Not expanded | 91.1 | 91 | 83.3 | 83.1 | 89 | 88 | 86.5 | 86.3 |
| ExD1, (WordNet) | 91.1 | 91.1 | 83.3 | 83.2 | 89.0 | 88.1 | 86.5 | 86.4 |
| ExD1, (WordNet + BERT Cos. Similarity ) | 91.3 | 91.2 | 83.5 | 83.3 | 89.2 | 88.2 | 86.7 | 86.5 |
| ExD2, (FastText) | 91.2 | 91.1 | 83.4 | 83.2 | 89.1 | 88.1 | 86.6 | 86.4 |
| ExD2, (FastText + BERT Cos. Similarity) | 91.4 | 91.2 | 83.6 | 83.3 | 89.3 | 88.2 | 86.8 | 86.5 |
| ExD4, M4 (BERT MASK) | 90.9 | 90.7 | 83.0 | 82.9 | 88.7 | 87.8 | 86.3 | 86.1 |
| ExD4, M4 (BERT MASK + BERT Cos. Similarity) | 91.6 | 91.5 | 83.6 | 83.4 | 89.3 | 88.3 | 86.8 | 86.6 |
| ExD, (Back-Translation) | 91.6 | 91.5 | 83.7 | 83.6 | 89.5 | 88.3 | 86.9 | 86.6 |
| ExD, (Back-Translation + BERT Cos. Similarity) | 91.7 | 91.6 | 83.9 | 83.7 | 89.6 | 88.6 | 87.1 | 86.9 |
| ExD, WordNet+ Fast-Text+ BT + BERTmask | 91.2 | 91.1 | 83.4 | 83.2 | 89.1 | 88.1 | 86.6 | 86.4 |
| ExD, GPT3 | **92.7** | **92.5** | **84.9** | **84.6** | **90.6** | **89.5** | **88.1** | **87.8** |
| ExD, WordNet+ Fast-Text+ BT + BERTmask + BERT Cos. Similarity | 92 | 91.7 | 84.2 | 83.8 | 89.9 | 88.7 | 87.4 | 87.0 |

Implementing BERT-cosine similarity as an additional filtering step consistently improved performance across all augmentation methods for each dataset. For example, when applied to the AskFm dataset, the enhancements were as follows: WordNet (0.2%), FastText (0.3%), Back-translation (0.6%), and BERT-masking (0.5%) compared to their counterparts without BERT-cosine similarity filtration. This indicates that BERT-cosine similarity effectively refines the augmented dataset by prioritizing semantically relevant augmentations. When all data augmentation techniques were combined, there was only a negligible increase in accuracy by 0.1% over the non-augmented dataset. However, integrating BERT-cosine similarity into the full augmentation pipeline led to a more significant accuracy improvement of 0.9% and an F1-score increase of 0.6%. This comprehensive approach, which incorporates multiple augmentation methods along with BERT-cosine similarity, highlights the synergistic potential of diverse augmentation strategies when effectively combined. Similar improvements were observed for the HASOC, TREC, and SST1 datasets as well. Nonetheless, without any modification, GPT-3 augmentation improved performance by accuracy 1.6% and 1.4% F1 score, outperforming all previous augmentation methods and our proposed method for enhancing augmentation techniques.

***Overfitting experiment for different augmentation methods*** Overfitting tends to be more severe when training on smaller datasets. By conducting experiments with a restricted fraction of the available training data, we demonstrate that BERT-cosine similarity filtration slightly improves classifier performance across different training set sizes by reducing overfitting effect as well. Our methodology involved comparing classifier performance across original training datasets, augmented training datasets (using FastText, WordNet, BERT-masking, and Back-translation), and datasets augmented then filtered by BERT-cosine similarity, at multiple augmentation levels: 1, 5, 10, and 20 times the size of the original dataset (Table 16). We adhered to a single-epoch training regime, consistent with our original experiments. For the AskFM dataset, our initial experiments without augmentation yielded a training F1 score of 94% and a validation F1 score of 92% on a 10k sample size. Post-augmentation, the total dataset size increased to nearly 364k. This approach was applied to multiple datasets, yielding similar observations. For instance, we observed an improvement in the training F1 score to 99.9% with up to 5 times augmentation of each sample, but only a marginal 0.1% improvement in the validation score, indicating negligible benefits in validation performance from traditional augmentation methods. However, after applying BERT-cosine similarity filtration, the dataset size was reduced to 60k, and we noted that after 5 times augmentation, the training F1 score was close to 99.9%, with improvement in validation F1 94.2%. This represents a 3% improvement over the baseline and a 2.9% improvement compared to the unfiltered, augmented dataset. Therefore, we conclude that traditional augmentation methods, which tend to produce data very similar to the original, do not provide additional learning benefits when used beyond 5 times sentence augmentation. A similar experiment with GPT-augmented sentences revealed significant improvements in both training and validation F1 scores—97% and 95.3%, respectively—up to 7 times augmentation of the original dataset size. Beyond this point, while training F1 scores continued to improve up to 99%, no further improvements were observed in validation scores. From this experiment, we infer that GPT-generated samples are effective up to 7 times the augmentation of the original dataset size, contrasting with traditional methods, which tend to overfit the model after 5 times augmentation.

**Table 16**
Overfitting experiment for different augmentation methods for AskFm dataset

| Augmentation Method | Level | Original Size | Augmented Size | Post-Filtration Size | Training F1 Score | Validation F1 Score (Baseline, Augmented, Post-filtrated) |
|---|---|---|---|---|---|---|
| None (Baseline) | 0x | 10k | - | - | 94% | 92% |
| Traditional | 1x, 5x, 10x, 20x | 10k | 20k 60k 100k 200k | 15k 17k 36k 70k | 97.2% 99.9% 99.9% 99.9% | 92% → 92.1% → 92.3% 92% → 92.1% → 94.1% 92% → 92.0% → 94.2% 92% → 92.1% → 94.2% |
| GPT-Augmented | 1x | 10k | 20k | - | 95% | 92% → 93.1% |
| GPT-Augmented | 5x | 10k | 50k | - | 97% | 92% → 95.3% |
| GPT-Augmented | 7x | 10k | 70k | - | 99.9% | 92% → 95.2% |

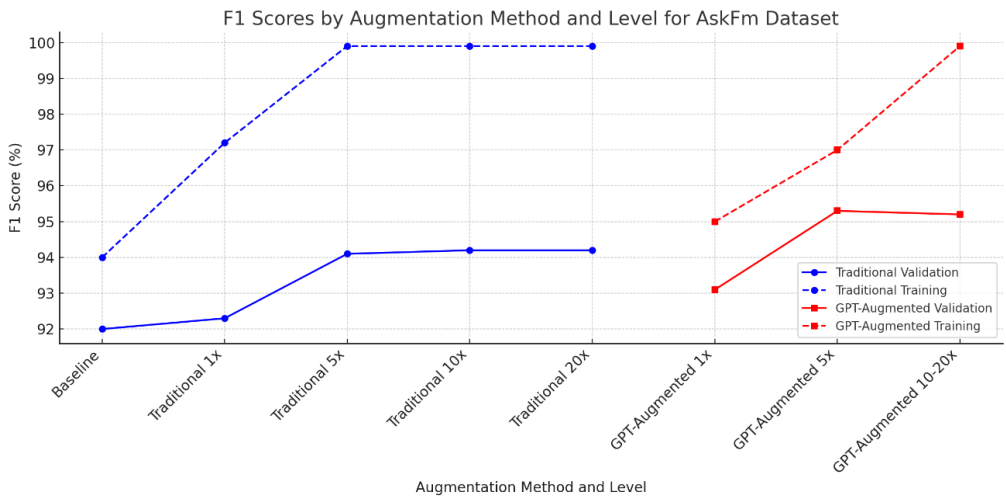

**Figure 12:** Comparative analysis of Training and Validation F1 scores across augmentation methods for the AskFm dataset.

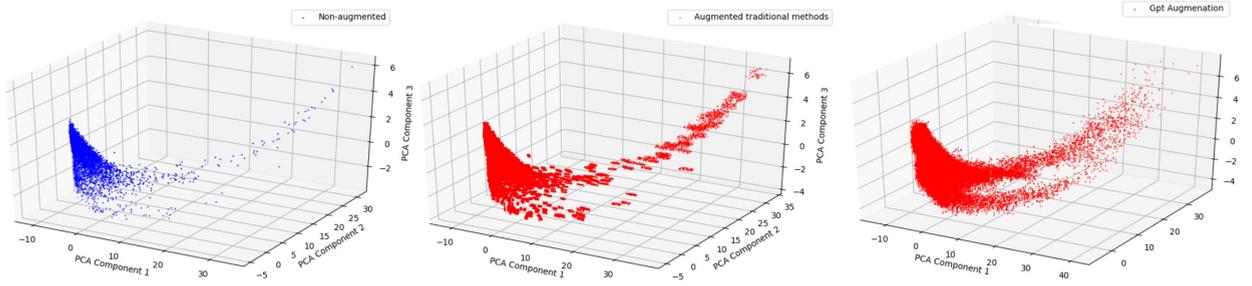

**Figure 13:** Comparative visualization of embedding space for non-augmented, traditional augmentation and GPT3 augmentation for the AskFm Dataset.

***Embedding Plots of Different Augmentation Methods*** To evaluate the impact of data augmentation on generalizability, we visualized each method's embeddings in a 3D plot using the last layer of BERT embeddings. Initially, we plotted embeddings for a non-augmented dataset of 10,000 samples, followed by visualizations of 360,000 samples augmented using four traditional methods (Figure 13). Notably, the density of the embeddings increased significantly due to the augmentation of similar sentences, with the augmented data being over 360 times larger. However, the overall area occupied by the augmented sentences did not expand significantly; most augmentations occurred within the boundaries of the original dataset, and only a 5% increase in area was observed. Using a filtered dataset after cosine filtration that was 100 times larger than the original, we observed similar patterns, which were almost indistinguishable by eye. However, after calculating area coverage, it was almost the same, with a 4.95% increase compared to non-augmented samples. This indicates that filtration might have no effect on generalizability, and at the same time, these traditional methods of augmentation lead to an increase in density but no significant expansion in the embedding space. This suggests that filtering augmented and non-augmented versions does not enhance coverage or might not result in generalized augmentation.

The same analysis was applied to a GPT-generated dataset, which was 20 times larger than the original dataset. Here, we observed an increase in the density of augmented sentences compared to the non-augmented dataset. Additionally, there was a notable 19.4% increase in the coverage area of augmentation embeddings, suggesting that GPT augmentation not only enhances sentence similarity but also expands the diversity of the dataset beyond that of traditional augmentation methods.

## 5. Error Analysis

After applying BERT cosine filtration, there was an improvement in accuracy and F1 score, although a small portion of false detection still persisted in the model. In order to gain insight into this phenomenon, we conducted a thorough analysis of the model's errors in this particular section. For this purpose, we randomly prepared 6 subsets of test data from the previous experiments (AskFm, and SST1 then manually inspected the classifier output. For manual inspection, each test data contained 100 samples. Our experimental setup and sample test data were identical for both non-expanded and expanded datasets. Since our results showed BERT outperformed others, we considered only BERT for error analysis. Table 17 presents error analysis results for both non-expanded and expanded datasets.

In AskFm dataset of 100 test samples, it contains 70 non-hate and 30 hate classes. With the non-expanded training model, overall accuracy was 91%, where 29 (98.57%) non-hate samples were classified correctly. However, only one non-hate sample was misclassified as hate which may have deceived the model since it has the word 'Dislike', which may trigger the model as a false negative. However, samples belonging to the hate classes were largely misclassified as non-hate. Before data augmentation, the models' success rate was 73.30% for hate samples. However, after data augmentation with BT, the accuracy of detecting hate samples improved to 76.60% and remain the same accuracy for non-hate detection. Finally, after data augmentation with BT+BERT Cos. Similarity, the accuracy of non-hate detection increased to 100% samples. However, still, some hate samples were misclassified as non-hate while it was initially labeled as hate. Among these samples, some words (i.e., thick, drive, Fuck) can have multiple meanings (sexual or non-sexual sense) based on the context. Besides, the word 'lamo', which is a social slang, is usually used for fun rather than serious conversation. For example, the sample sentence 'lmao when u studpid fuck' was misclassified as a non-hate, and this might look as a straight error. However, if we further examine the issue, we may notice that the post has several elements that might deceive the classifier (e.g., stupid wrongly spelled as 'studpid', 'u' used as short form of 'you', and 'lamo' often used for fun). Similar improvement detection has been observed in confusion matrix analysis as well. For example, we have drawn a confusion matrix of 500 test samples, where 350 non-hate samples and 150 hate samples. We have seen that the best performance was observed after the BERT-cosine application, where accuracy has improved for all classes for both AskFm, and SST1 dataset (Fig 14 and 15).

**Table 17**
Error analysis of BERT before and after dataset augmentation. Accuracy (%) False detection determined by manual checking of 100 posts from each of four datasets. Classifier results are obtained from BERT architecture.

| Dataset | Non-hate detection Acc. (%) | Hate detection Acc. (%) | Overall Acc. | Mis-classified sentences and improvements |
|---|---|---|---|---|
| **AskFm Not-Expanded** | 98.57 | 73.30 | 91 | **One non-hate sample misclassified as non-hate, e.g.:**<br>1. What would someone have to do to make you dislike them immediately?<br>**Eight hate samples misclassified as non-hate, e.g.:**<br>1. this person is 100 fake so beware Isabella<br>2. No Offense of any sort is intended in this question. Homosexuals, do you find being with your non preferable sex gross, just like guys find it gross to be with other guys?<br>3. no he a bitch<br>4. eyes but i here you have a nice butt to how about driving from back<br>5. how short is your shortest shorts i head its small<br>6. lmao when u studpid fuck |
| AskFm Expanded BT | 98.57 | 76.60 | 92 | - One previously miss-classified hate sentence (no. 3) has rightly classified after expanded. |
| AskFm ExD,+BT + BERT Cos. Similarity | 100 | 76.60 | 93 | - One previously miss-classified non-hate sentence (no. 1) has rightly classified after expanded. |

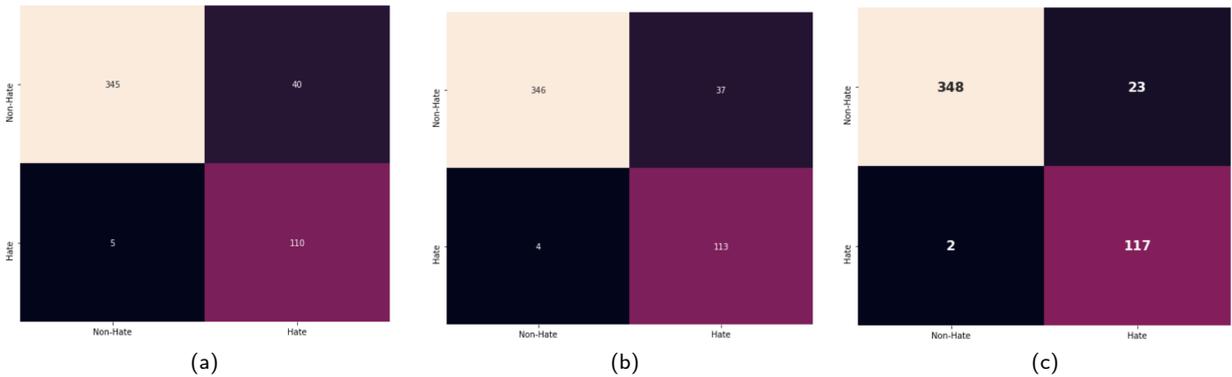

(a)                                                (b)                                                (c)

**Figure 14:** Confusion matrix of detection of hate and non-hate samples for AskFm dataset. (a). the non-augmented dataset, (b). augmented dataset using Back-translation (BT), (c). augmented dataset using BT + BERT cos. similarity filtration. Here BERT is used for classifier performance, and a total of 500 test samples have been used.

## 6. Discussion

The above experiments highlighted several vital findings for NLP data augmentation. We utilized four prominent data augmentation techniques. Our analysis reveals that each method has its advantages and disadvantages in terms of augmented dataset size, diversity of augmentation, risk of label alteration, and the impact on classifier performance. For instance, BT demonstrated minimal label alteration and significant improvement in the F1 score but was limited by generating a smaller number of augmented sentences. Conversely, WordNet, FastText, and BERT-MLM were capable of producing a larger volume of augmented sentences; however, they also introduced a higher rate of label alterations, potentially undermining reliability for supervised learning. Table 18 presents a summary of our findings, drawing comparisons from previous results in Tables 10, 12, and 15. These tables detail the outcomes of different data augmentation methods in terms of dataset size, label alteration risk, and classifier performance before and after applying BERT cosine similarity to the data augmentation methods. We observed that BERT cosine similarity effectively filtered incorrectly annotated sentences, enhancing performance compared to traditional augmentation methods. However, during tests for overfitting, traditional data augmentation methods tended to overfit after augmenting a sentence five times, whereas GPT-3 allowed for sevenfold augmentation of each sentence without overfitting. Traditional methods also fell short in terms of generalizability; for instance, the average area under the curve (AUC) increased by 15% less following augmentation compared to GPT-3's augmentation. Although applying BERT cosine similarity resulted in a 0.7% average improvement in the F1 score and almost 0% label alteration, it did not significantly enhance generalizability when compared to GPT-3. Although we only utilized a baseline version of GPT-3, its performance could potentially be further enhanced by exploring methods like Retrieval-Augmented Generation (RAG) and Lanchain, which we have not yet tested due to the initial improvements observed with just GPT-3 augmentation. Future research could investigate the effects of data augmentation on subcategory labels, which we did not explore in this experiment despite observing improvements across several domain datasets. A more nuanced experiment could involve more complex texts and targeting subcategory labels of augmentation to compare performance. Additionally, employing more complex datasets for augmentation could allow for the observation of the hallucination effect in Large Language Models (LLMs).

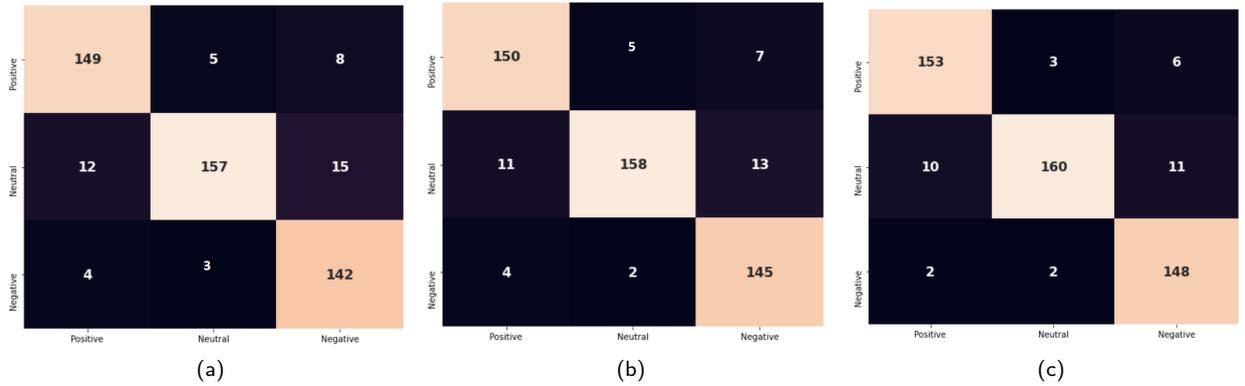

**Figure 15:** Confusion matrix of detection of Positive, Neutral and Negative samples for SST1 dataset. (a). non-augmented dataset, (b). augmented dataset using Back-translation (BT), (c). Augmented dataset using BT + BERT cos. similarity filtration. Here BERT is used for classifier performance and total 495 test sample been used.

## 6.1. Comparison with Related Work

Most studies previously explore data augmentation as a complementary result for a task-specific context, so it is hard to directly compare our DA with previous literature. Moreover, negligible work has experimented with different DA in combination with other methods; also, experiments with different domains were highly overlooked.

But there are some studies similar to ours that evaluate augmentation techniques on multiple datasets. Nevertheless, none of the work can be compared precisely with ours, even though they have used the same dataset in some cases. The reason is that they have used different ML models, and sometimes training and testing samples were different. In this circumstance, though the exact result might not be compared, however, we can correlate our overall findings to theirs and get a comparative insight.

A work by [42] implemented easy augmentation methods which used synonym replacement insertion, deletion, and swap of a random word and applied to SST2, TREC, and MR dataset and observed improvement of classifier accuracy. Their experiment showed EDA yields an improvement in accuracy claimed minor label alteration. However,

**Table 18**

Performance summary of different augmentation methods before and after BERT-cosign similarity applied. Results of average from Table 10 (datasets augmentation size), 12 (percentage of label alteration), and 15 (classifier performance improvement). Label alteration was determined by manually checking 500 posts from each of the four datasets. Classifier results are obtained from BERT architecture.

| Method | Augmented dataset size | % of label alteration | Classification F1 score improvement |
|--------|------------------------|------------------------|-------------------------------------|
| Before BERT-cosine similarity applied | | | |
| WordNet | 88 times | 2.6% | 0.1% |
| FastText | 95 times | 1.9% | 0.1% |
| Back-Translation | 2.9 times | .87% | 0.4% |
| BERT Mask | 42 times | 3.18% | - 0.4% |
| WordNet+ FastText+ BT + BERT mask | 150 times | 0.1 | 0.1 |
| After BERT-cosine similarity applied to data augmentation methods. | | | |
| WordNet+BERT Cos. Similarity | 70.5 times | 0.1% | 0.2% |
| FastText+BERT Cos. Similarity | 40.2 times | 0% | 0.2% |
| Back-Translation +BERT Cos. Similarity | 2 times | 0% | 0.5% |
| BERT Mask BERT Cos. Similarity | 42 times | .25% | 0.6% |
| WordNet+ FastText+ BT + BERT mask + BERT Cos. Similarity | 80 times | **0.05%** | **0.7%** |
| GPT3 | 20 times | **0.01%** | **1.4%** |

this work presents average classier improvement for all datasets used; therefore, it was unclear which dataset improved EDA methods. Furthermore, this work does not present separate results for each EDA method, making it impossible to know which EDA methods were suitable. Their average improvement of EDA methods was 1.6% of accuracy. If we compare this work with ours, we have mixed opinions which agree and disagree with their findings. For example, we used synonym replacement with wordNet and FastText as EDA methods, and our finding shows that yes, EDA as synonym replacement improves classifier performance; however, it was not applicable for all domains. For instance, the TREC dataset showed a decrease in performance after synonym replacement. Also, we have a dispute that EDA does not alter labels much, unlike their findings. Our findings showed that EDA method label alteration were high in a range of 3-5% depending on different domain datasets.

Another work, [2] has worked with six hate speech domain datasets, including AskFm and FromSpring dataset like ours, and used Back-translation and achieved 5% accuracy improvement. Though that experiment does not consider the risk of label alteration, our experiment repeats that Back-translation indeed improves classifier accuracy and F1 scores for all datasets with the minimum label alteration.

Furthermore, [22] has worked with BERT-based contextual augmentation using different pre-trained BERT models and used BERT as the classifier. Their experiment showed an improvement in classifier performance after BERT contextual augmentation for SST2 and TREC datasets. In our case, we have also seen BERT-contextual augmentation improved accuracy and F1 scores for hate speech and sentiment domains but was not suitable for all datasets. For example, the TREC dataset, which was text categorization, showed a drop of classifier performance 0.2% ( 86.3% to 86.1%) after BERT-mask data augmentation. In this regard, we can affirm partially with their results that BERT contextual DA improves outcomes of the classifier performance. However, since our experiments yield the highest number of label alterations for BERT DA and have not enhanced classifier improvement for all datasets, therefore, we dispute that BERT-contextual DA is suitable for all domain datasets. In addition, their work does not provide any insight into the risk of label alteration or error analysis while using BERT as DA. Therefore it was not possible to comment BERT DA risk of label alteration.

## 7. Conclusion

We presented four popular data augmentation practices in this study, including simple synonym replacement (Word-Net, FastText), Back-translation, and contextual synonym replacement (BERT-mask). We have tested all of these previous augmentation techniques with five benchmarked labeled datasets. Our goal was to find out which augmentation method works better. Initially, we have seen none of the methods (contextual or simple synonym replacement) can satisfactorily generate sentences without label alteration that are close to the original sentence. After manual checking of 500 augmented sentences from each dataset, WordNet and BERT-mask augmented sentences showed higher label alteration in the range of 4-6%. In contrast, back-translation showed the lowest label alteration, 0.3-1.5%. However back-translation method produced fewer sentences compared to other methods. The BERT-based contextual synonym replacement produces the highest number of sentences with a wide diversity; however, it still produces more than 6% of sentences with altered labels. To alleviate this problem, we have proposed a BERT-based contextual cosine similarity and filtered out sentences that have lower contextual similarity than original sentences. The result of our experiment was promising since it has reduced label alteration to only 0.05% and worked for all experimented domains.

Subsequently, we evaluated the augmented datasets using both traditional and our proposed methods with various ML classifiers, including CNN and BERT base. Our approach showed an improvement of an average of 0.7% in classification accuracy over the other methods and exhibited the least amount of label alteration. However, while our method was effective for traditional augmentation, it encountered limitations in terms of generalizability and susceptibility to overfitting. Specifically, traditional augmentation techniques failed to show promise, leading to model overfitting after a 5-fold increase in data augmentation. In contrast, GPT-3 was able to augment data up to seven times without overfitting, and it demonstrated a 15% larger area coverage in the embedding space compared to traditional approaches. Furthermore, sentences augmented with GPT-3 showed a 1.4% improvement in ML classification experiments over traditional methods and a 0.8% improvement over our proposed methods. Our findings indicate that, traditional augmentation techniques offer marginal performance gains, augmenting data with Large Language Models (LLMs) like GPT-3 can significantly enhance machine learning model performance.

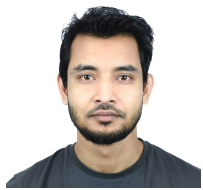

**Md Saroar Jahan:** received his BSc in electrical engineering from the Department of Electrical Engineering Electronics, United International University, Bangladesh, in 2013, and his Msc degree from University of Oulu in 2020 in Computer Science and Engineering. He is currently pursuing a Ph.D. degree in Computer Science at the University of Oulu, Finland. His research interests include Big data analysis, deep learning techniques, and Natural Language processing.

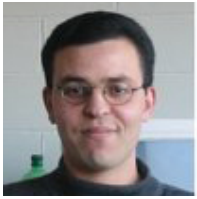

**Mourad Oussalah:** received his MSc in Control engineering form University of Paris XII France in 1994 and PhD degree in Robotics/Computer Science in Evry Val Essonnes University in France. After academics positions in KU Leuven, City University of London and University of Birmingham, he is since 2016 with University of Oulu as a Research Professor leading the Social Mining Research Group. His research focuses on data mining and uncertainty handling where he published more than 220 papers and led several projects in the field.